\documentclass[letterpaper, 10 pt, conference]{ieeeconf}  

\IEEEoverridecommandlockouts                              

\usepackage{amsmath,amsfonts}
\usepackage{algorithmic}
\usepackage{algorithm}
\usepackage{array}
\usepackage[caption=false,font=normalsize,labelfont=sf,textfont=sf]{subfig}
\usepackage{textcomp}
\usepackage{stfloats}
\usepackage{url}
\usepackage{verbatim}
\usepackage{ifpdf}
\ifpdf
  \usepackage{graphicx}
  \usepackage{hyperref}
\else
  \usepackage[dvipdfmx]{graphicx}
  \usepackage[dvipdfmx]{hyperref}
\fi
\usepackage{bm}
\usepackage{here}
\usepackage{comment}
\usepackage{cite}
\usepackage{color}

\title{\LARGE \bf
Control Design for a Rideable Animatronic Two-Wheeled Robot \\ with Quadruped Form
}

\author{Tadashi Sumioka$^{1}$, Kazushi Akimoto$^{1}$,  Takuya Tsujimura$^{1}$, Sho Takayanagi$^{1}$ and Yuya Horiuchi$^{2}$\\
\thanks{*This work was not supported by any organization}
\thanks{$^{1}$Tadashi Sumioka, Kazushi Akimoto, Takuya Tsujimura, Sho takayanagi and Tomoyo Konya are with Honda R\&D Co., Ltd., Innovative Research Excellence, Frontier Robotics, Saitama 3510114, Japan
        {\tt\small tadashi\_sumioka@jp.honda}
        }%
\thanks{$^{2}$Yuya Horiuchi is with Honda Motor Co., Ltd. }
}

\begin{document}

\maketitle
\thispagestyle{empty}
\pagestyle{empty}

\begin{abstract}
In recent years, motorcycle popularity has been
declining, particularly among the younger generations. To rekindle
interest in motorcycles among this demographic, we developed
a rideable two-wheeled robot equipped with four limbs as a
future partner mobility concept and intended for use in public
events. The character reproduced by this robot carries the
protagonist on its back and exhibits a dynamic quadrupedal
gait. To maximize the riding experience, we aimed to match the
robot's weight and size to the character's specifications while
ensuring rider safety and enabling expressive movements of
limbs, wrists, ankles, and facial features. However, achieving locomotion
solely through limb movement would require excessive
motor output and increased limb strength, resulting in higher
weight and extremely slow gait, thereby reducing character
fidelity. To overcome these challenges, the robot performs its
primary locomotion using a self-balancing two-wheeled base,
while the limbs provide auxiliary support during mounting
and dismounting and move in coordination with the wheeled
locomotion speed. This approach enables an animatronic robot
capable of intuitive, weight-shift-based control for free and
natural movement. In this paper, we focus on a robust selfbalancing
control method that maintains stability even during
rapid limb movements, as well as a motion control strategy that
generates natural quadruped-like behavior.
\end{abstract}

\begin{keywords}
Two-wheeled mobile robot, Animatronics, Human-Centered Robotics, Human-Robot Collaboration, Self-Balancing Control.
\end{keywords}

\section{Introduction}
\par In recent years, motorcycle use has been declining among the younger generations in developed countries, a trend that is particularly pronounced in urban areas \cite{yamamoto}. This shift is largely driven by improvements in the convenience of automobiles and public transportation, as well as growing concerns about safety and environmental impact \cite{suzuki}. Nevertheless, motorcycles remain a unique form of mobility that offers a highly immersive experience, allowing riders to dynamically engage their entire body and feel a strong sense of unity with the machine. To share this experiential value and rekindle interest among young people, we developed a rideable robot that faithfully reproduces a popular character\footnote[3]{\copyright Pok$\acute{{\mathrm{e}}}$mon. \copyright Nintendo/Creatures Inc. /GAME FREAK inc. TM, \textregistered, and character names are trademarks of Nintendo.}. The character being replicated carries the protagonist on its back and exhibits dynamic quadrupedal gaits. Our design prioritizes accurate reproduction of the character's size and appearance, leveraging animatronics technology to create a safe partner mobility platform that preserves the immersive qualities of motorcycle riding.
\begin{figure}[tb]
\begin{center}
\includegraphics[width=80mm]{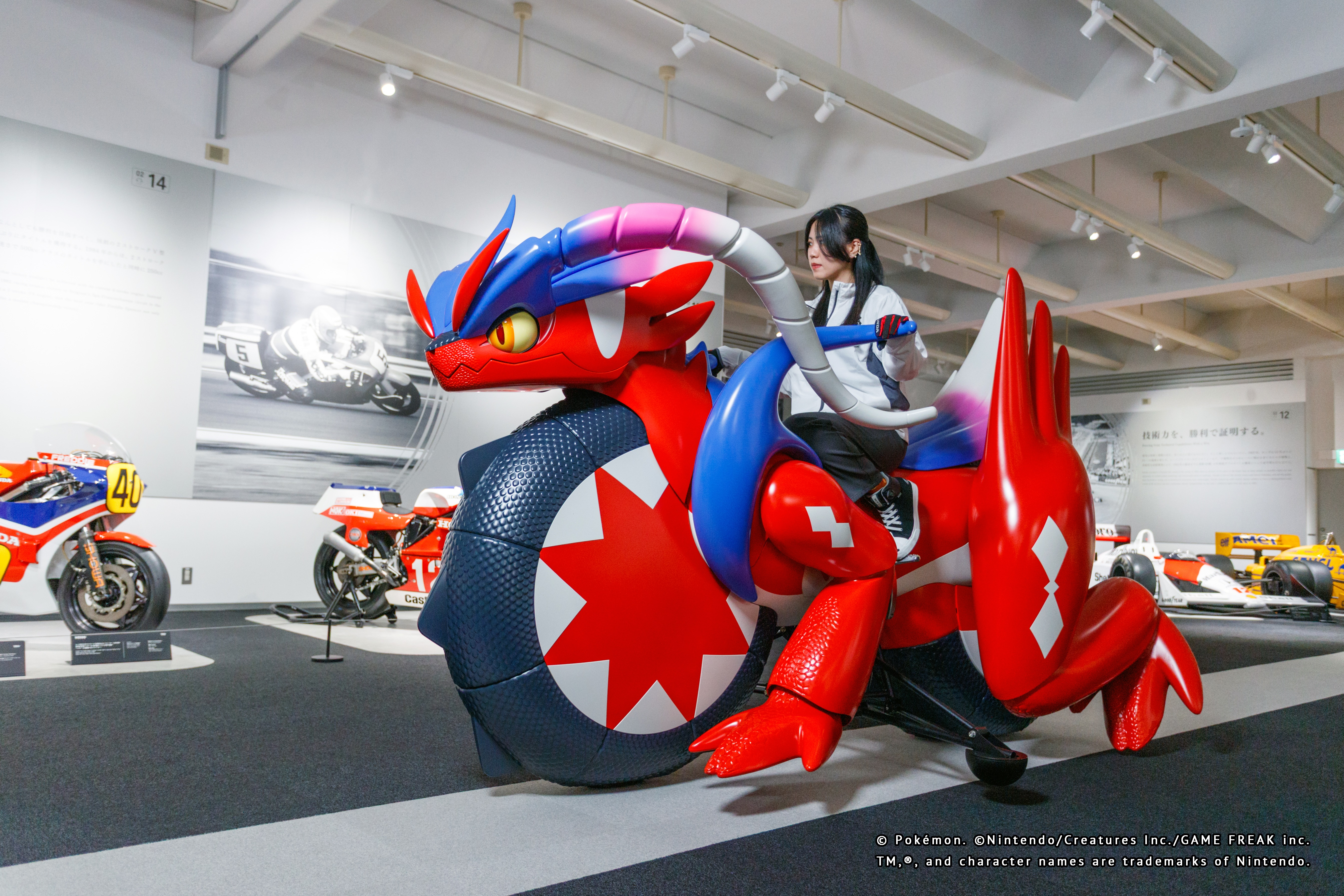}
\caption{Photograph of our proposed future partner mobility.}
\label{fig:riding_assist_vehicle}
\end{center}
\end{figure} 
\par Animatronics combines physical motion with visual and emotional expression to enhance user interaction, and its applications have expanded in education and entertainment (\cite{DR1}, \cite{DR2}, \cite{DR3}, \cite{hirose}, \cite{johnson}). Recent advances using reinforcement learning have achieved remarkable results in generating lifelike motions for animatronic systems that balance physical stability with expressive character behavior (\cite{DR1}, \cite{DR2}, \cite{DR3}, \cite{unitree}). However, when attempting to maintain character fidelity while enabling locomotion through active limb propulsion, we found that motor output and structural strength constraints made this approach impractical.
\par To address this, the robot's primary locomotion is realized through a two-wheeled base, while the limbs assist during mounting and dismounting and perform synchronized cyclic motions during travel without generating thrust. Additional expressive movements of the limbs, wrists, ankles, and head were incorporated to enhance the sense of dynamism.
\par To maximize the immersive riding experience, the robot's weight and dimensions match the original character specifications. Rideable animatronic robots are rare, and our system introduces a novel form of human–machine interaction that integrates expressive appearance with functional mobility. Unlike conventional animatronics, our platform operates as a Human-in-the-Loop (HiL) system, where the rider's actions directly influence control. Because simulating the full range of rider behaviors is infeasible, reinforcement learning with Sim-to-Real(S2R) transfer cannot guarantee stable locomotion. Furthermore, rider safety must take precedence, making machine-learning-based safety assurance impractical at present.
\par Character fidelity also imposes structural constraints: the seat position is higher than that of typical motorcycles, preventing riders from placing their feet on the ground, and the steering system cannot replicate conventional motorcycle geometry. Consequently, self-stabilization through natural steering dynamics is minimal, requiring autonomous balance control both at standstill and during wheeled locomotion.
\par Given these challenges, the core technical requirement is a self-stabilizing control method that enables a heavy, high-center-of-gravity two-wheeled structure with actively moving multi-DOF manipulators to remain stable and maneuverable. Conventional two-wheeled vehicles are inherently sensitive to center-of-gravity shifts and external disturbances, and this complexity is amplified in rideable robots with dynamic upper structures (\cite{kondo}, \cite{tanaka}).
\par To achieve this, our system integrates disturbance observers and a Virtual External Force(VEF) framework—techniques proven in bipedal robot control and automotive stability systems (\cite{takenaka_VEF}, \cite{toyoshima}). These methods allow stable and responsive maneuvering even when the limbs and neck undergo large movements, preserving character expressiveness without compromising ride stability.
\par Beyond animatronics, such control technologies have potential applications in disaster-response robotics and next-generation mobility systems. This paper presents the control approach and outdoor riding results for a two-wheeled animatronic robot with actively moving limbs and neck, operated by a rider using intuitive weight-shift.

\section{Overview of Control Design and Hardware Configuration}
\begin{figure*}[tb]
\begin{center}
\includegraphics[width=140mm]{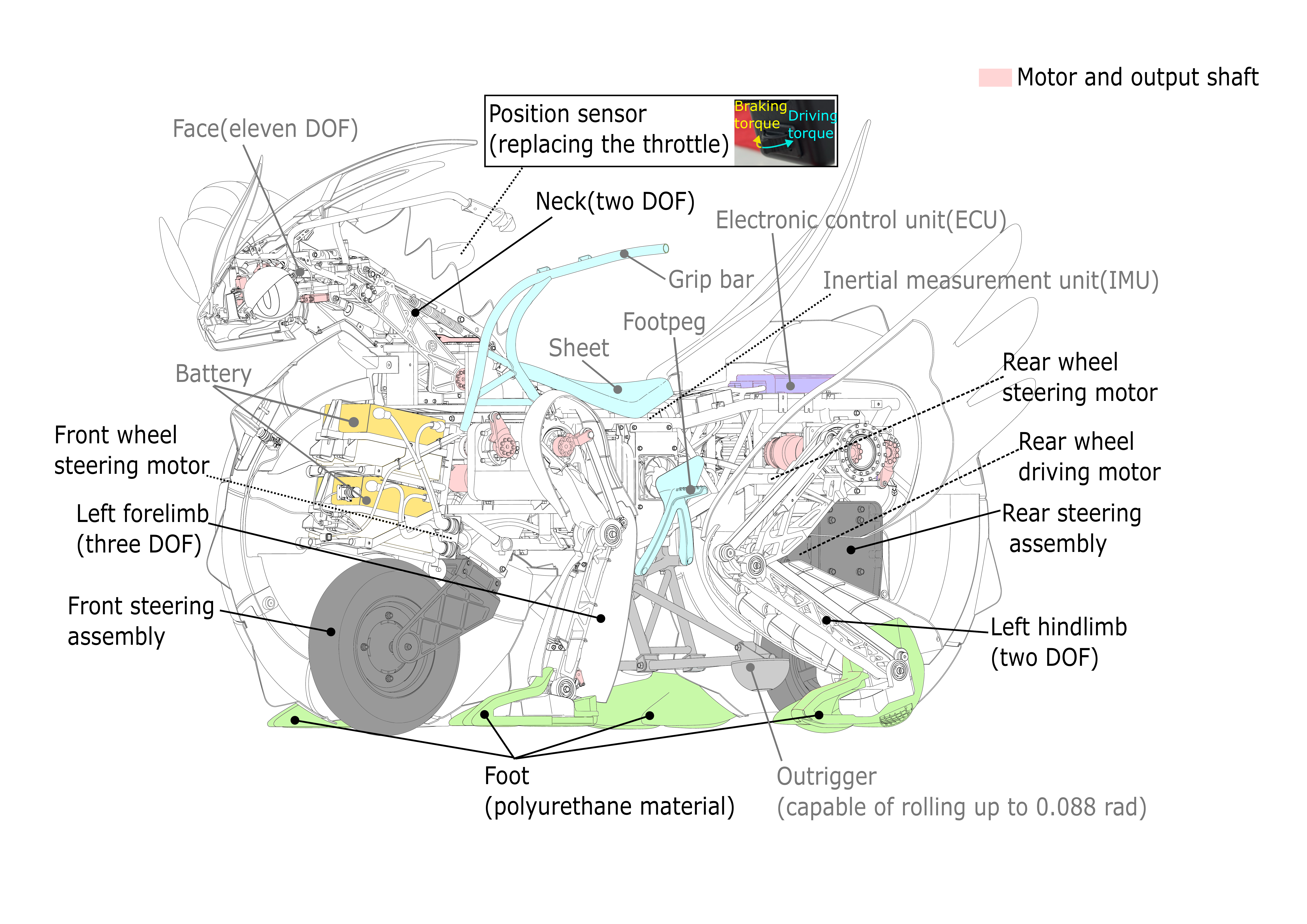}
\caption{Mechanical design of our robot. The robot has three degrees of freedom per forelimb and two degrees of freedom per hindlimb. The neck has two degrees of freedom. The angles of each motor, the rear wheel velocity, the posture of the robot body, and the pressing force applied to the grip bar are measured and utilized for control. Driving torque is generated by pulling the position sensor attached to the grip bar, while braking torque is generated by pushing it.}
\label{fig:mecha_des}
\end{center}
\end{figure*}

\begin{figure*}[tb]
\begin{center}
\includegraphics[width=170mm]{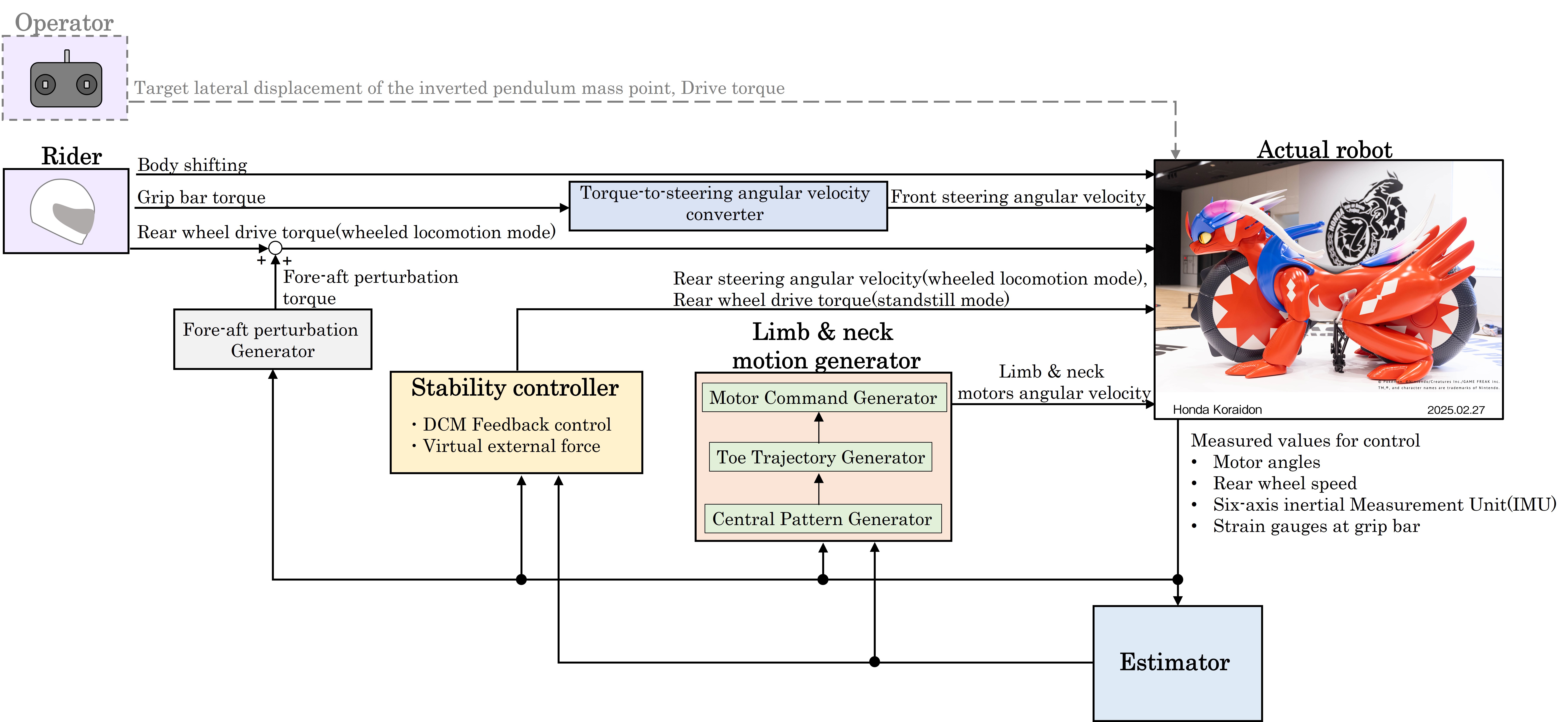}
\caption{Overall diagram of the control scheme.}
\label{fig:control_scheme}
\end{center}
\end{figure*}

\begin{figure*}[tb]
\begin{center}
\includegraphics[width=160mm]{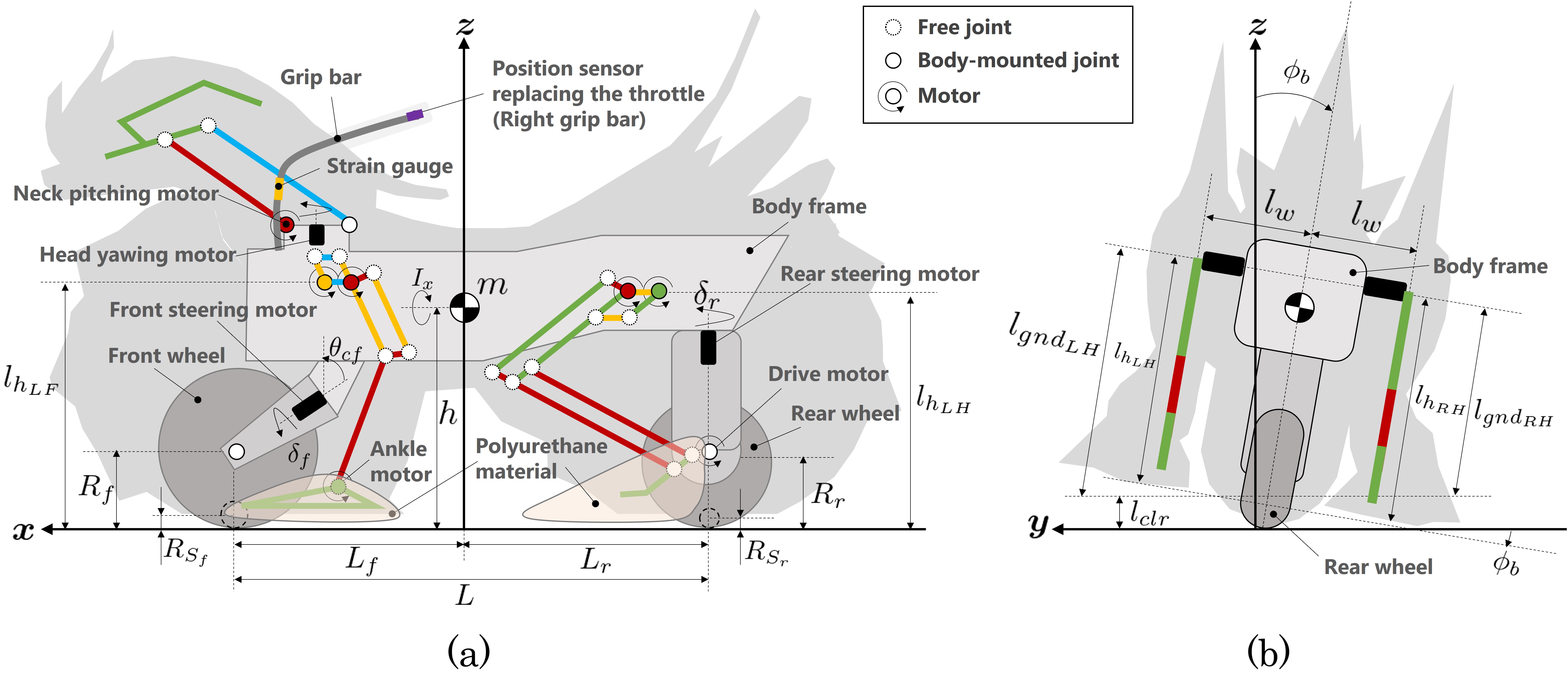}
\caption{The hardware configuration, the definition of variables and the coordinate system used in this study. (a) Left side view. (b) Rear view. The front limbs are equipped with three motors, while the hind limbs are equipped with two. Two motors are implemented in the neck and head to generate yaw and pitch motions. The robot is equipped with steering motors for both the front and rear wheels.}
\label{fig:hardware}
\end{center}
\end{figure*}

\begin{figure*}[tb]
\begin{center}
\includegraphics[width=160mm]{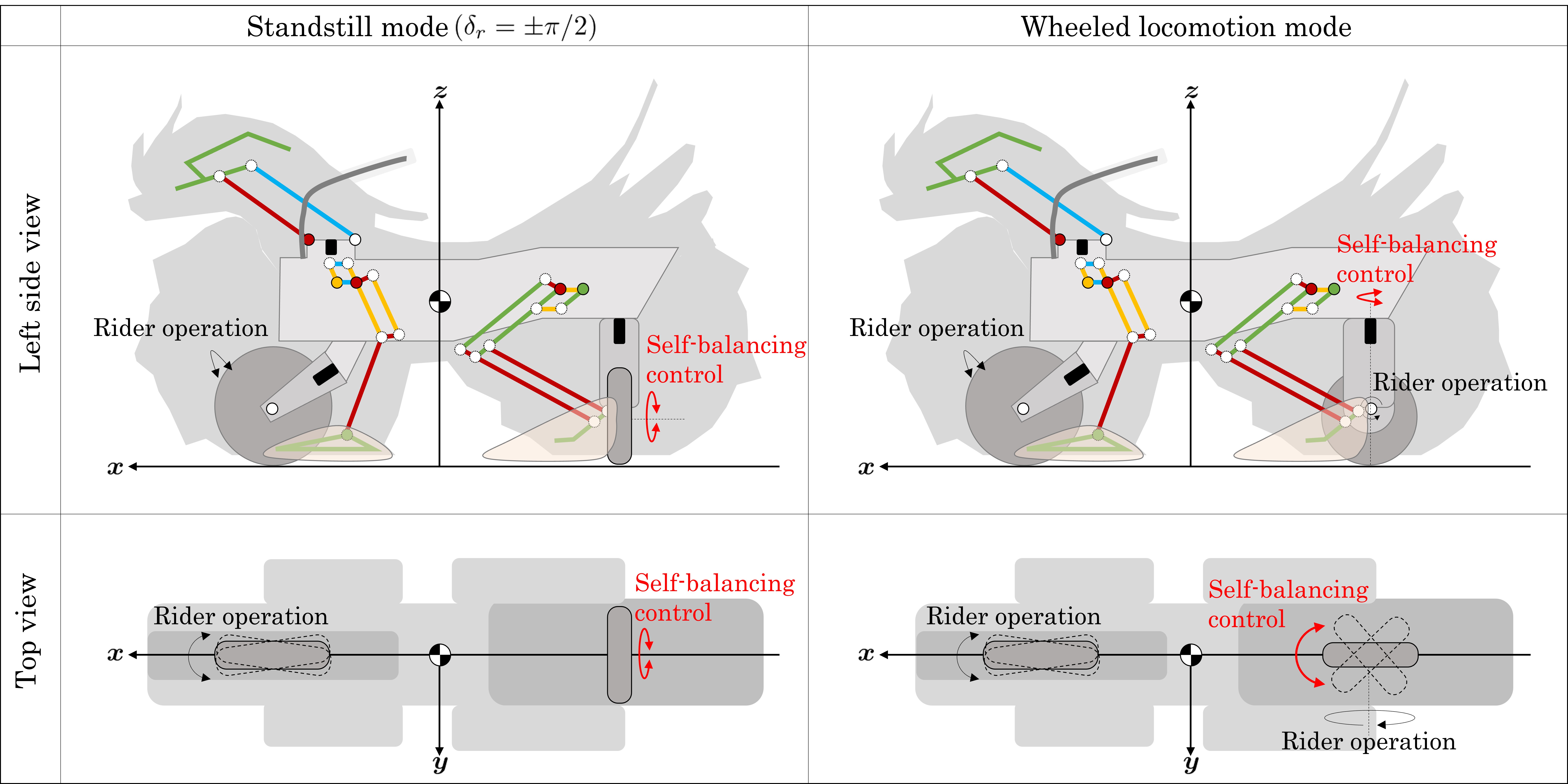}
\caption{Standstill mode and wheeled locomotion mode in self balancing control}
\label{fig:control_mode}
\end{center}
\end{figure*}

The design of the robot developed in this study is shown in Fig.~\ref{fig:mecha_des}. The forelimbs and hindlimbs possess three and two degrees of freedom in the sagittal plane, respectively, while the neck has two degrees of freedom. The feet are made of polyurethane material, allowing them to bend naturally upon contact with the ground. In addition, the rider's footpegs were positioned so as not to interfere with the limbs and to facilitate ease of mounting.
The control architecture designed in this study is shown in Fig.~\ref{fig:control_scheme}.
The robot's locomotion is realized using a self-balancing two-wheeled vehicle, while the limbs and neck are coordinated to move in accordance with its wheeled locomotion speed. Consequently, a robust and easily operable stabilization control method, together with a motion control strategy that evokes locomotion using limbs similar to the original character, becomes essential. If the coupling between the limb and neck motions is misaligned, or if their respective motions fail to match the wheeled locomotion speed, significant incongruity arises, undermining the validity of the system as an animatronic robot. To address these challenges, we employed a Central Pattern Generator (CPG), which governs rhythmic motor patterns in biological systems, to synchronize the overall motion. Furthermore, maintaining a constant wheeled locomotion speed would impart a mechanical impression and reduce the fidelity of character reproduction; therefore, we incorporated fore-aft perturbation, a pulsation observed in the forward direction of the original character and in walking organisms. In addition, to enable demonstrations without a rider, the system is designed to be operable via a wireless controller.
The symbols used in this paper are defined in Table \ref{table:variables} and Fig.~\ref{fig:hardware}\footnote[4]{$i \in \{ f, r \}$, $f$ denotes the front wheel and $r$ denotes the rear wheel.}. 
\begin{table}[hbtp]
  \caption{Definitions of Variables}
  \label{table:variables}
  \centering
  \begin{tabular}{llll}
    \hline
    Symbol  & Units &Description  \\
    \hline \hline
    $m $  & kg  & Total mass \\
    $h $  & m  & Height of center of gravity \\
    $I_{x} $  & kgm$^{2}$  & Total roll inertia about center of gravity \\
    $g $  & m/s$^{2}$  & Gravitational acceleration \\
    $\delta_{i} $  & rad  & Steering angle \\
    $R_{i} $  & m  & Wheel radius \\
    $R_{S_{i}} $  & m  & Wheel sectional radius \\
    $\theta_{ci} $  & rad  & Caster angle \\
    $L $  & m  & Wheel base \\
    $L_{i} $  & m  & Length from wheel axle to center of gravity \\
    $\phi_{b}$  & rad  & Roll angle of vehicle body (lean angle) \\
    $\omega_{z}$  & rad/s  & Yaw rate of vehicle body \\
    $\omega_{i}$  & rad/s  & Wheel rolling angular velocity \\
    $V_{ox}$  & m/s  & Logitudinal velocity of center of gravity \\
    $V_{oy}$  & m/s  & Lateral velocity of center of gravity\\
    $\tau_{drive}$  & Nm  & Rear wheel driving torque \\
    \hline
  \end{tabular}
\end{table}
If the front and rear wheels were sized similarly to those used in conventional motorcycles, they would interfere with the exterior design, making it difficult to faithfully reproduce the original character. Therefore, the front and rear wheels of the robot were designed to be relatively small. The front wheel was given a limited steering angle range to avoid interference with the exterior.
The rear wheel was fixed at $\theta_{cr} = 0$ and allowed to steer within the range $\delta_{r} \in [-\pi/2,\ \pi/2]$, enabling balance assistance primarily through rear-wheel control. As shown in Fig.~\ref{fig:control_mode}, during standstill, the rear wheel is set to $\delta_{r} = \pm \pi/2$ and actively driven to maintain balance. During wheeled locomotion, the rider controls the rear wheel drive while rear-wheel steering is used to stabilize the robot. 
For front-wheel steering, strain gauges attached to the grip bar detect the rider's steering torque, which is then used to generate the front-wheel steering angle. This design allows the rider to intuitively follow the desired trajectory. 
\par An overview of the control scheme developed in this study is shown in Fig.~\ref{fig:control_scheme}. 

In the following sections, the subscript $d$ is used to denote desired values for various variables. The index $j$ represents each limb, where $j \in \{ LF, RF, LH, RH \}$ corresponds to Left Forelimb, Right Forelimb, Left Hindlimb, and Right Hindlimb, respectively. The subsequent chapters describe the detailed control methods for the limbs and neck, as well as the balance control strategy within the proposed control scheme.

\section{Limb \& Neck motion}
\subsection{Rhythm Generation Using a Central Pattern Generator}
\par The Central Pattern Generator (CPG) is a neural circuit that generates rhythmic movement patterns in biological systems, and has been applied in robotic gait control \cite{CPG1} and rhythmic walking assist devices \cite{CPG2}. In this paper, the phase is computed as follows:
\begin{eqnarray}
\theta_{CPG} = \int \frac{\pi}{w_{traj}}V_{ox} \ dt. \label{theta_CPG}
\end{eqnarray}
Here, $w_{traj}$ represents the toe displacement during the stance phase, as described later in Fig.~\ref{fig:limb_link}. In the robot developed in this study, $w_{traj}$ differs between the forelimbs and hindlimbs. Therefore, the term $\pi/w_{traj}$ on the right-hand side of equation (\ref{theta_CPG}) was set to 3.5 based on the visual appearance of the robot during actual wheeled locomotion. 
As shown in Fig.~\ref{fig:limb_link}, since $w_{traj}$ is larger for the hind limbs than for the fore limbs, reducing the value of $\pi/w_{traj}$ delays the lift-off timing of the hind limbs, resulting in a tendency for increased dragging of the hind limbs.
It is known that quadrupedal animals change their gait depending on their locomotion speed \cite{quadrupedal1}. To reproduce both trot and gallop gaits as seen in the original character, we determined the phase offset $\Delta \theta_{j}$ for each limb according to the wheeled locomotion speed, as shown in Table \ref{table:limb_Delta}, and applied it to the phase in equation (\ref{theta_CPG}). 
\begin{table*}[hbtp]
  \caption{Setting of Phase Differences for Each Limb}
  \label{table:limb_Delta}
  \centering
  \begin{tabular}{cccc}
    \hline
      & Stop($V_{ox} < $ $\varepsilon_{v}$) & Trot($ \varepsilon_{v} \leq V_{ox} \leq v_{th}$)  & Gallop($V_{ox} > v_{th}$)   \\
    \hline \hline
    $ \Delta \theta_{LF} $ & 0 & 0 & $\pi$  \\
    $ \Delta \theta_{RF} $ & 0 & $\pi$ & $\pi$ \\
    $ \Delta \theta_{LH} $ & 0 & $\pi$ & 0\  \\
    $ \Delta \theta_{RH} $ & 0 & 0 & 0 \\
    \hline
  \end{tabular}
\end{table*}
In Table \ref{table:limb_Delta}, $\varepsilon_{v}$ represents a small velocity threshold used to determine a standstill state, and $v_{th}$ denotes the speed threshold at which the gait transitions from trot to gallop.
Based on these definitions, the desired phase value for each limb is computed as follows: 
\begin{eqnarray}
\theta_{j_{d}} = \theta_{CPG} + \Delta \theta_{j}. \label{theta_d}
\end{eqnarray}
As shown in Table \ref{table:limb_Delta}, $\Delta \theta_{j}$ changes stepwise according to the wheeled locomotion velocity. Consequently, the desired phase value $\theta_{j_{d}}$ in Equation (\ref{theta_d}) also changes in a stepwise manner. Therefore, the actual phase $\theta_{j}$ for each limb is computed using the following first-order lag model:
\begin{eqnarray}
\dot{\theta}_{j} = K_{\theta}\left(\theta_{j} - \theta_{j_{d}} \right). 
\end{eqnarray}
Here, $K_{\theta}$ is a negative constant. By setting the absolute value of this constant to a sufficiently small value, abrupt changes in the actual phase $\theta_{j}$ can be avoided even when the desired phase $\theta_{j_{d}}$ changes stepwise. 
 On the other hand, if the absolute value of $K_{\theta}$ is too small, the convergence of the limbs is delayed during deceleration from wheeled locomotion to a stop, exacerbating air stepping. 
Based on experimental results with the actual robot, $K_{\theta}$ was set to $-0.5$. 
\subsection{Gait generation}
\subsubsection{Toe trajectory}
\begin{figure*}[tb]
\begin{center}
\includegraphics[width=140mm]{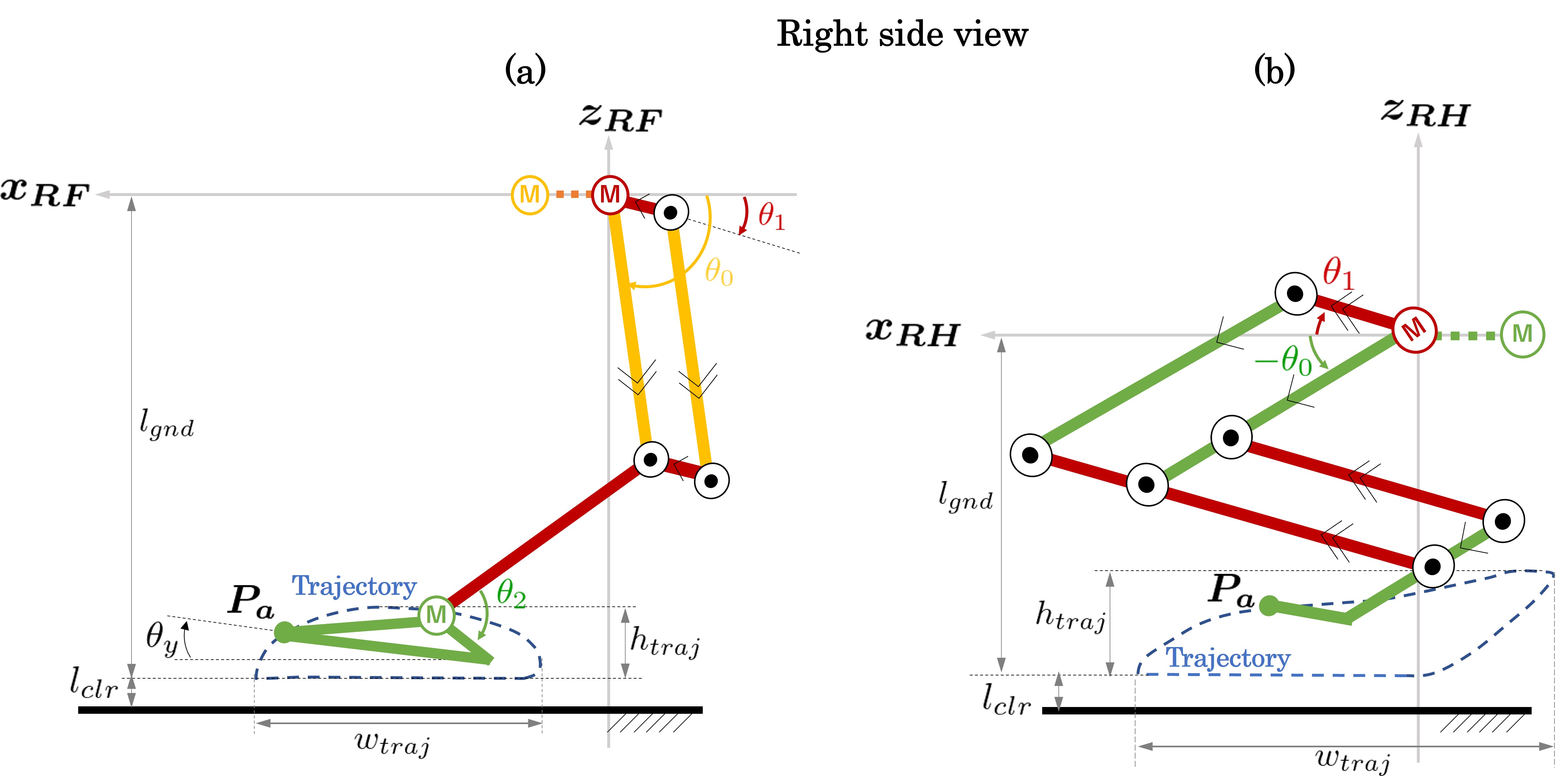}
\caption{Details of the limb linkage mechanism.  (a) Right forelimb, with the shoulder joint as the origin. (b) Right hindlimb, with the hip joint as the origin. $\theta_{y}$ denotes the pitch angle of the front foot sole.}
\label{fig:limb_link}
\end{center}
\end{figure*}

\begin{figure}[tb]
\begin{center}
\includegraphics[width=100mm]{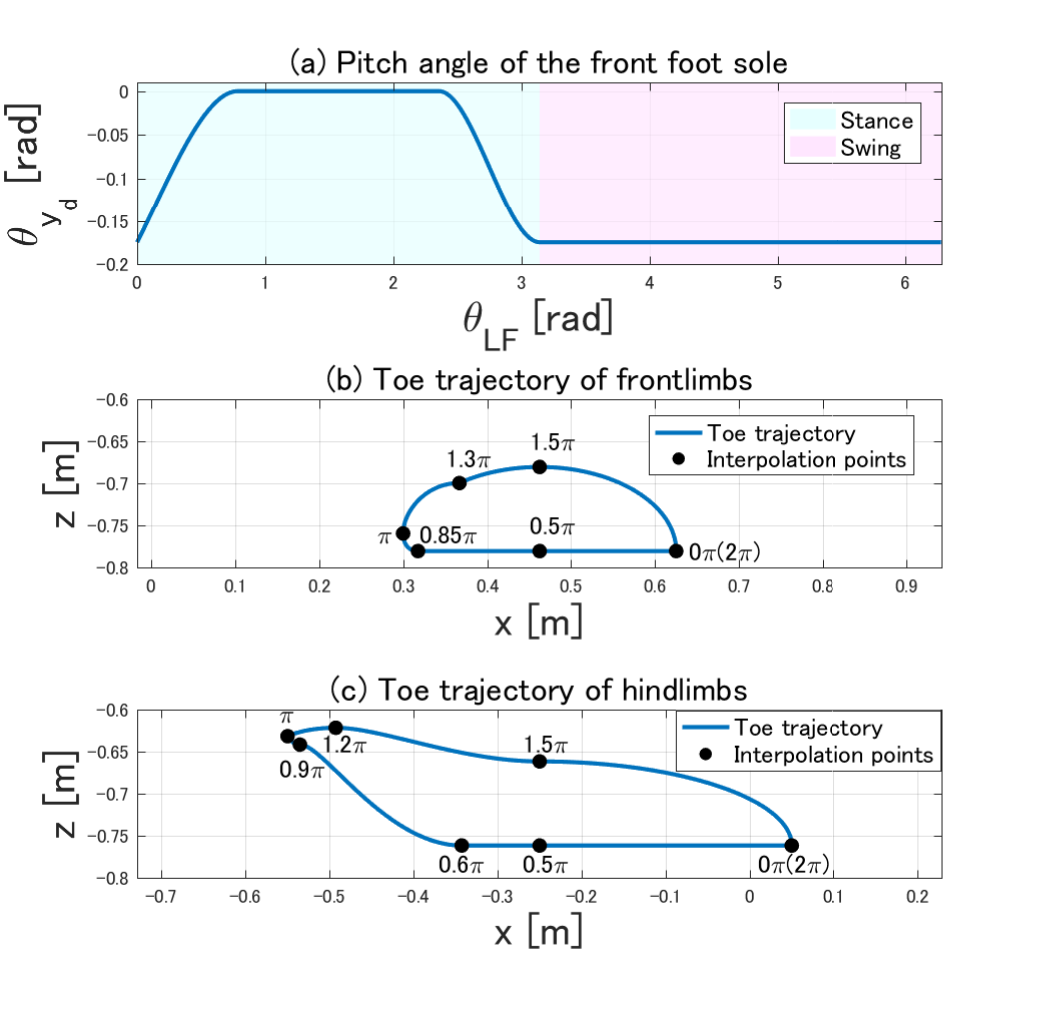}
\caption{Desired position and orientation of the ${\bm{P_{a}}}$. The swing-phase trajectory is designed based on a sine wave, with interpolation points added to generate the desired path.}
\label{fig:toe_tra}
\end{center}
\end{figure}

As shown in Fig.~\ref{fig:limb_link}, gait generation is achieved by making the point of action of the terminal link, denoted as ${\bm{P_{a}}}$ follow a desired trajectory. In the original character, even on flat ground, the forelimbs exhibit a toe-first contact gait. Therefore, as illustrated in Fig.~\ref{fig:limb_link}(b), the forelimbs are equipped with three motors, and a desired posture $\theta_{y_{d}}$ is also defined.
Specifically, considering the range of motion constraints imposed by interference with the exterior components, the forelimbs are controlled to follow the desired posture and trajectory shown in Figure \ref{fig:toe_tra}. 

As noted in \cite{toe_traj1} and \cite{toe_traj2}, the toe trajectory of animals that walk with a heel-first contact tends to have a higher initial swing-phase elevation. Therefore, the hindlimb trajectory was designed as shown in Fig.~\ref{fig:toe_tra}(c).
To ensure tracking of the trajectory shown in Fig.~\ref{fig:toe_tra}, the angular velocity commands for each motor, $\dot{\theta}_{0}, \dot{\theta}_{1}, \dot{\theta}_{2}$ were derived using resolved-rate control based on the Levenberg-Marquardt method\cite{sugihara_IK}. 

\subsubsection{Ground Clearance Calculation Considering Robot Body Roll Angle}
During the stance phase, we consider computing a trajectory that maintains a constant clearance from the ground as if the limb were actively pushing against it, even when the robot body roll angle changes. Let $l_{gnd_{j}}$ denote the distance from the shoulder or hip joint to the desired ground clearance. As shown in Fig.~\ref{fig:hardware}(b), this can be geometrically calculated as follows: 
\begin{eqnarray}
l_{gnd_{LF}} &=& l_{h_{LF}} + l_{w}\tan\phi_{b} + l_{clr}\cos\phi_{b}, \label{l_gnd_LF} \\
l_{gnd_{RF}} &=& l_{h_{RF}} - l_{w}\tan\phi_{b} + l_{clr}\cos\phi_{b}, \label{l_gnd_RF} \\
l_{gnd_{LH}} &=& l_{h_{LH}} + l_{w}\tan\phi_{b} + l_{clr}\cos\phi_{b}, \label{l_gnd_LH} \\
l_{gnd_{RH}} &=& l_{h_{RH}} - l_{w}\tan\phi_{b} + l_{clr}\cos\phi_{b}. \label{l_gnd_RH} 
\end{eqnarray}

Here, on the actual robot, the setting was $l_{clr} = -0.05$ m, which corresponds to the foot sole being lifted 0.05 m above the ground.

\subsection{Neck pitch motion}
Using $\theta_{CPG}$ obtained from equation (\ref{theta_CPG}), the pitch angle of the neck is determined as follows: 
\begin{eqnarray}
\theta_{neck\_{pitch}} = -G_{neck}\sin\theta_{CPG}. 
\end{eqnarray}
Here, it is assumed that the origin of the neck pitch motion is set at $\theta_{neck\_{pitch}} = 0$. The parameter $G_{neck}$ represents the amplitude, which was set to 0.035 rad on the actual robot based on the allowable range of motion constrained by the robot’s exterior structure. 

\subsection{Fore-aft Perturbative Motion}
\begin{figure}[tb]
\begin{center}
\includegraphics[width=80mm]{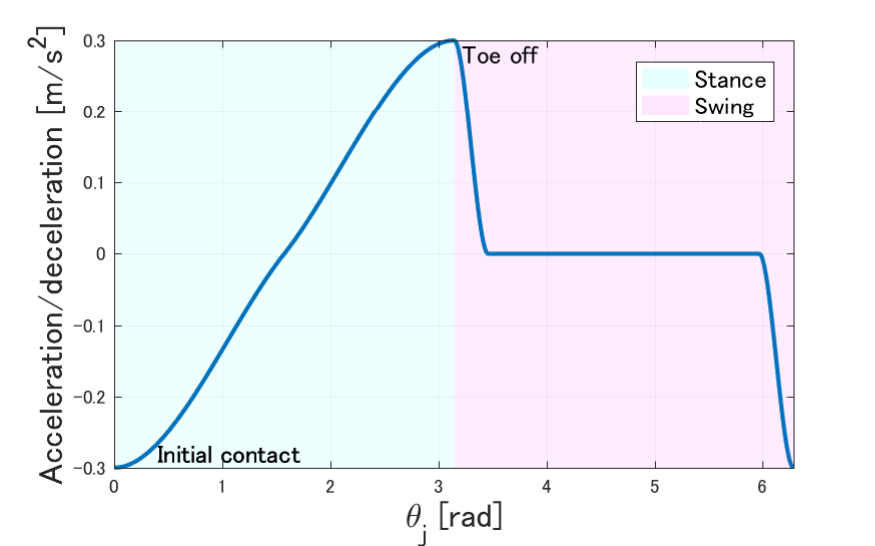}
\caption{Acceleration/deceleration during stance phase.}
\label{fig:rocking}
\end{center}
\end{figure}

\begin{figure}[tb]
\begin{center}
\includegraphics[width=80mm]{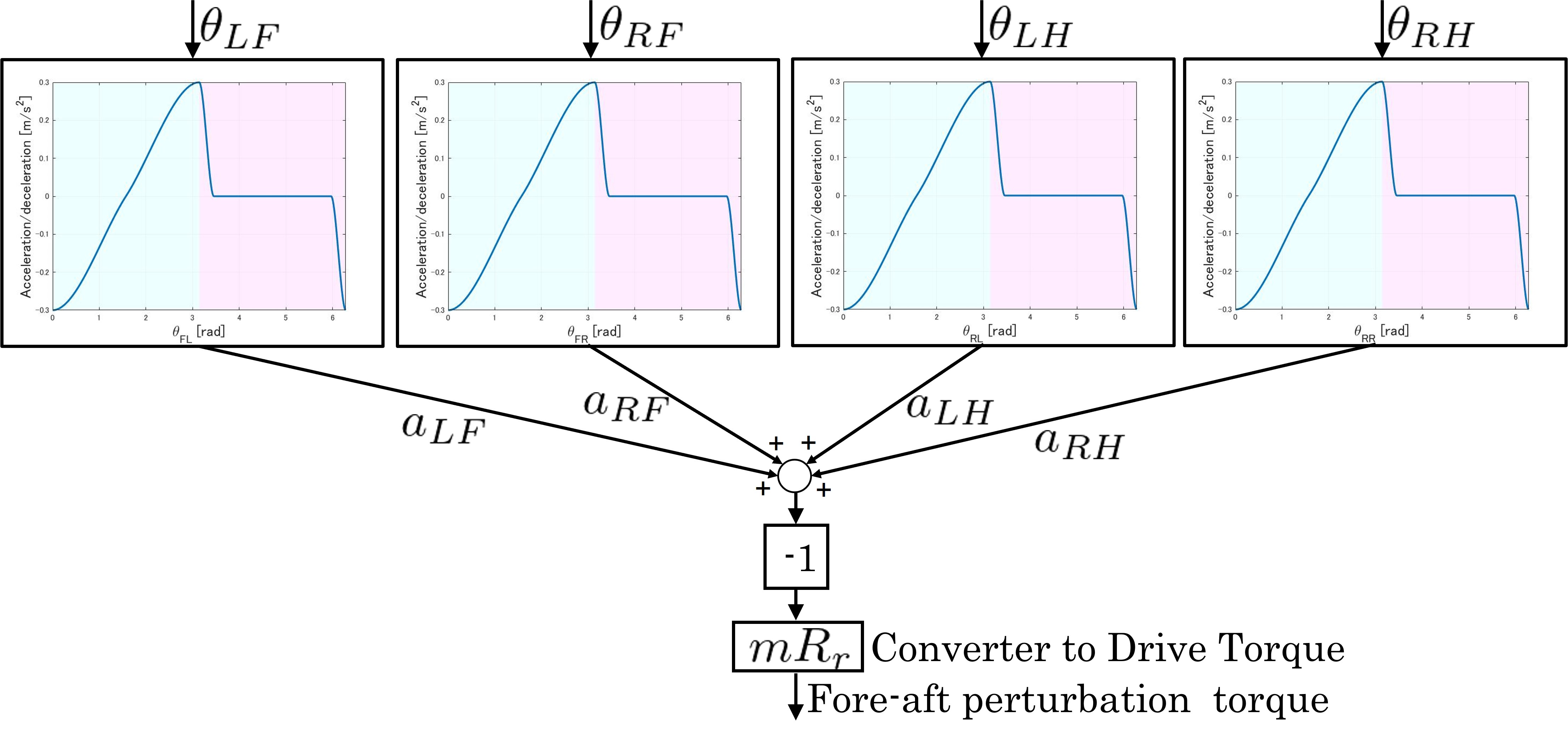}
\caption{Fore-aft perturbation torque generator}
\label{fig:fore_aft}
\end{center}
\end{figure}

According to \cite{Winter}, it has been shown that the anterior-posterior ground reaction force transitions as follows during the stance phase. 
\begin{itemize}
\item At the initial stage of the stance phase, a backward force is generated from the perspective of the direction of progression, in order to absorb the impact of ground contact and decelerate the body.
\item In the late stage of the stance phase, a forward force is generated in the direction of progression to produce propulsive a force that accelerates the body through push-off.
\end{itemize}
Because of the occurrence of acceleration and deceleration as described above, the anterior-posterior velocity of the center of mass in walking animals exhibits slight perturbations. Therefore, referring to the waveform presented in \cite{Winter}, acceleration and deceleration command values are generated as shown in Fig.~\ref{fig:rocking}, based on the phase of each limb. The total acceleration/deceleration value of the robot body, as illustrated in Fig.~\ref{fig:fore_aft}, is obtained by summing the contributions from all four limbs and converting this into a drive torque, which is then added to the torque commanded by the rider to realize fore-aft perturbation. It should be noted that although the limbs exhibit acceleration and deceleration as shown in Fig.~\ref{fig:rocking}, the robot body, which serves as the base of the limb links, must reproduce the opposite behavior relative to the limbs, thus requiring a final multiplication by -1. By computing the required acceleration and deceleration from the phase of each limb in this manner, fore-aft perturbation can be reproduced using the same computational method, even when the gait pattern differs, such as in a trot or a gallop. 

\section{Stability control}
Since the control laws differ between the standstill and wheeled locomotion states, stability control is implemented by separating modes using a state machine. In both control modes, the control gains are configured to realize divergent component of motion (DCM: \cite{takenaka_1}- \cite{englsberger_2}) feedback control, which is known to be the best stabilizing regulator for an inverted pendulum under input constraints \cite{sugihara} and its stability and effectiveness, in the sense that it can maximize the stability margin, have also been demonstrated in two-wheeled systems \cite{sumioka}. 
\subsection{Standstill mode}
The controller is constructed based on the control model proposed in \cite{sumioka}. The equivalent two-point mass decomposition linear inverted pendulum(E2PMD-LIP) model introduced in \cite{sumioka} is as follows:
\begin{eqnarray}
\frac{d}{dt}
\left[ 
\begin{array}{c} 
P_{1y} \\ 
\dot{P}_{1y}
\end{array} 
\right ]
=
\left[ 
\begin{array}{cc} 
0 & 1 \\ 
\omega^2 & 0
\end{array} 
\right ]
\left[ 
\begin{array}{c} 
P_{1y} \\ 
\dot{P}_{1y}
\end{array} 
\right ]
+
\left[ 
\begin{array}{c} 
0 \\ 
-1
\end{array} 
\right ]
u.  \label{state_eq}
\end{eqnarray}
Here, $P_{1y}$ denotes the lateral displacement of the E2PMD-LIP point mass, $\omega$ represents the natural frequency of the E2PMD-LIP model, and $u$ is the lateral acceleration input applied to the E2PMD-LIP point mass. To implement DCM feedback control, $u$ is determined according to equation (\ref{u_DCM}):
\begin{eqnarray}
u &=& K_{\xi}\left(P_{1y} - P_{1yd} \right) + \frac{K_{\xi}}{\omega}\left(\dot{P}_{1y} - \dot{P}_{1yd} \right). \label{u_DCM} 
\end{eqnarray}
Here, $K_{\xi}$ denotes the feedback gain. 
$P_{1yd}$ represents the desired value of $P_{1y}$ transmitted when the robot is operated via a wireless controller without a rider. In the presence of a rider, the rider controls the desired position through weight shifting; therefore, $P_{1yd}$ is set to zero. Additionally, $\dot{P}_{1yd}$ is also set to zero. 
Assuming that $\delta_{f} = 0$, $\delta_{r} = \pm \pi/2$, and $V_{ox} = 0$ during standstill balancing, the following relationship holds between the rear wheel stabilizing drive torque $\tau_{drive}$ and the input $u$:
\begin{eqnarray}
mhu = \tau_{drive}. \label{tau_drive}
\end{eqnarray}
From equations (\ref{u_DCM}) and (\ref{tau_drive}), the rear-wheel-drive torque $\tau_{drive}$ during standstill balancing can be obtained as follows:
\begin{eqnarray}
\tau_{drive} = mh\left( K_{\xi}\left(P_{1y} - P_{1yd} \right) + \frac{K_{\xi}}{\omega}\dot{P}_{1y} \right)\sin\delta_{r}. 
\end{eqnarray}
Note that since $\delta_{r} = \pm \pi/2$, the term $\sin\delta_{r}$ is multiplied to adjust the sign accordingly. 
\subsection{Wheeled locomotion mode}
\subsubsection{Base model}
When attempting to implement self balancing control during wheeled locomotion using the control law proposed in \cite{sumioka}, a challenge emerged: even slight inclinations of the road surface triggered rear-wheel steering to maintain balance, resulting in continuous rear steering input and difficulty in maintaining straight-line motion. To prevent the rear-wheel steering assist from responding too sensitively, the following model is defined\footnote{Since $\delta_{f}$ can only take small values, it is excluded from the model for simplicity and instead considered within the disturbance observer structure presented later.}:
\begin{eqnarray}
\frac{d}{dt}
\begin{bmatrix}
P_{1y} \\
V_{by} \\
V_{ry} \\
\end{bmatrix}
=
\begin{bmatrix}
0 & 1 & -\frac{L_{f}}{L} \\
\omega^2 & 0 & \frac{V_{ox}}{L} \\
0 & 0 & 0 \\
\end{bmatrix}
\begin{bmatrix}
P_{1y} \\
V_{by} \\
V_{ry} \\
\end{bmatrix}
+
\begin{bmatrix}
0 \\
0 \\
1 \\
\end{bmatrix}
\dot{V}_{ry}. \label{state_eq_run}
\end{eqnarray}
Here, $V_{ry}$ denotes the lateral velocity of the rear wheel and is expressed as follows:
\begin{eqnarray}
V_{ry} = R_{r}\omega_{r}\sin\delta_{r} \label{V_ry}.
\end{eqnarray}
In addition, $V_{by}$ represents the following value, which is the sum of the lateral velocity generated by the roll motion of the E2PMD-LIP and the lateral velocity at the center of gravity induced by steering during driving: 
\begin{eqnarray}
V_{by} = \dot{P}_{1y} + V_{oy} \simeq \dot{P}_{1y} + \frac{L_{f}}{L}V_{ry}. \label{V_by}
\end{eqnarray}
As shown in (\ref{state_eq_run}), we consider minimizing $V_{ry}$ by constructing an augmented system that includes $V_{ry}$ as a state variable and applying state feedback control. Specifically, we consider the following state feedback control using $\dot{V}_{ry}$ as the control input:
\begin{eqnarray}
u = \dot{V}_{ry} = 
\begin{bmatrix}
K_{P} & K_{V} & K_{ry} \\
\end{bmatrix}
\begin{bmatrix}
P_{1y} - P_{1yd} \\
V_{by} - V_{byd} \\
V_{ry} \label{u_run} \\
\end{bmatrix}.
\end{eqnarray}
When the desired poles for the above feedback control are specified as $\lambda_{1}, \lambda_{2}, \lambda_{3}$, and $\lambda_{3}$ is set to $-\omega$ to construct a DCM feedback control, the gains $K_{P}, K_{V}, K_{ry}$ are given as follows:
\begin{eqnarray}
K_{P} &=& \frac{L\omega\left( \omega - \lambda_{1} \right)\left( \omega - \lambda_{2} \right)}{L_{f}\omega - V_{ox}}, \label{K_P} \\
K_{V} &=& \frac{K_{P}}{\omega}, \\
K_{ry} &=& \lambda_{1} + \lambda_{2} - \omega .
\end{eqnarray}
Here, $P_{1_{y_{d}}}$ represents the desired value of $P_{1_{y}}$ transmitted from the wireless controller during remote operation. When a rider is on board, $P_{1_{y_{d}}} = 0$. Additionally, $V_{by_{d}}$ denotes the desired value of $V_{by}$ determined by the VEF loop, which is designed to achieve both disturbance suppression and rider maneuverability. From (\ref{V_ry}), $\dot{V}_{ry}$ is given as follows:
\begin{eqnarray}
\dot{V}_{ry} = R_{r}\omega_{r}\dot{\delta}_{r}\cos\delta_{r} + R_{r}\dot{\omega}_{r}\sin\delta_{r}.
\end{eqnarray}
Therefore, the rear-wheel steering angular velocity command value  is given as follows:
\begin{eqnarray}
\dot{\delta}_{r} = \frac{u - R_{r}\dot{\omega}_{r}\sin\delta_{r}}{R_{r}\omega_{r}\cos\delta_{r}}.
\end{eqnarray}
\subsubsection{Enhancing both robustness and maneuverability using virtual external force}
\begin{figure}[tb]
\begin{center}
\includegraphics[width=80mm]{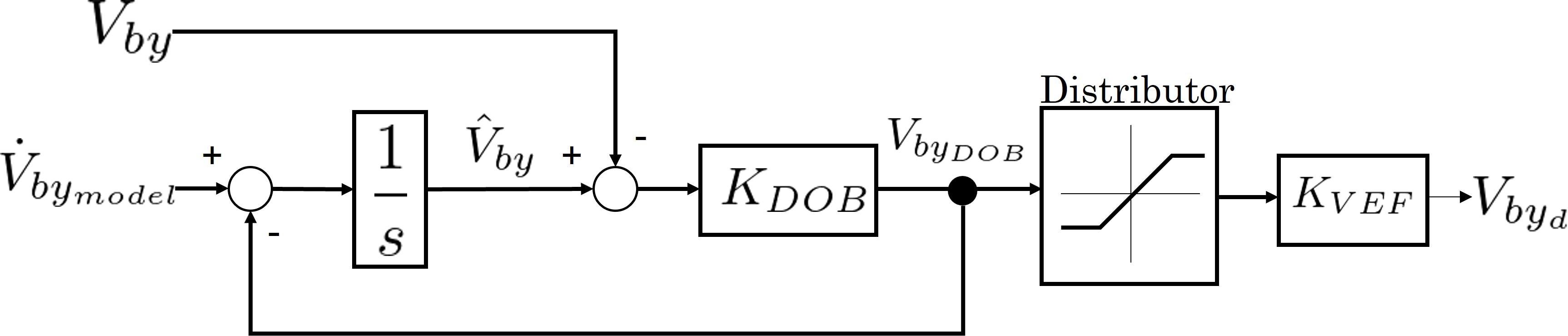}
\caption{Desired value generation of $V_{by}$ using virtual external force.}
\label{fig:DOB}
\end{center}
\end{figure}

The use of only the previously described control scheme to generate coordinated motion of the limbs and neck often led to instability, particularly at the onset of wheeled locomotion, which prevented the robot from achieving self-balancing. 
To achieve both disturbance suppression caused by limb and neck movements and rider maneuverability, the VEF framework was introduced into the system. The VEF framework is a method proposed by Takenaka for stabilizing control of bipedal walking robots \cite{takenaka_VEF}, and its application to vehicle stability control and maneuverability enhancement has also been reported \cite{toyoshima}. The VEF framework generates the desired state variables using an internal model, and the term “virtual external force” refers to the feedback component that  asymptotically drives the measured state variables toward the desired state variables. By employing this framework, the desired state variables adapt to the rider's control inputs, thereby improving maneuverability. Furthermore, when an observer model is used in the internal model, the structure becomes equivalent to a disturbance observer, which enhances robustness against disturbances. In implementing the state feedback control described in (\ref{u_run}), the desired value $P_{1_{y_{d}}}$ is provided by the wireless controller as previously mentioned. Therefore, the VEF loop is constructed using a minimal-order observer for $V_{by}$, as described in (\ref{V_by_DOB}) and illustrated in Fig.~\ref{fig:DOB}:
\begin{eqnarray}
\dot{V}_{by_{DOB}} = -K_{DOB}V_{by_{DOB}} - K_{DOB}\left( \dot{V}_{by} - \dot{V}_{by_{model}} \right). \label{V_by_DOB}
\end{eqnarray}
Here, $K_{DOB}$ and $K_{VEF}$ are positive gains. Furthermore, $\dot{V}_{by{model}}$ is given as follows:
\begin{eqnarray}
\dot{V}_{by_{model}} = \omega^2 \left( P_{1y} - P_{1yd} \right) + \frac{V_{ox}}{L}V_{ry}.
\end{eqnarray}
\section{Result}
\begin{figure*}[tb]
\begin{center}
\includegraphics[width=140mm]{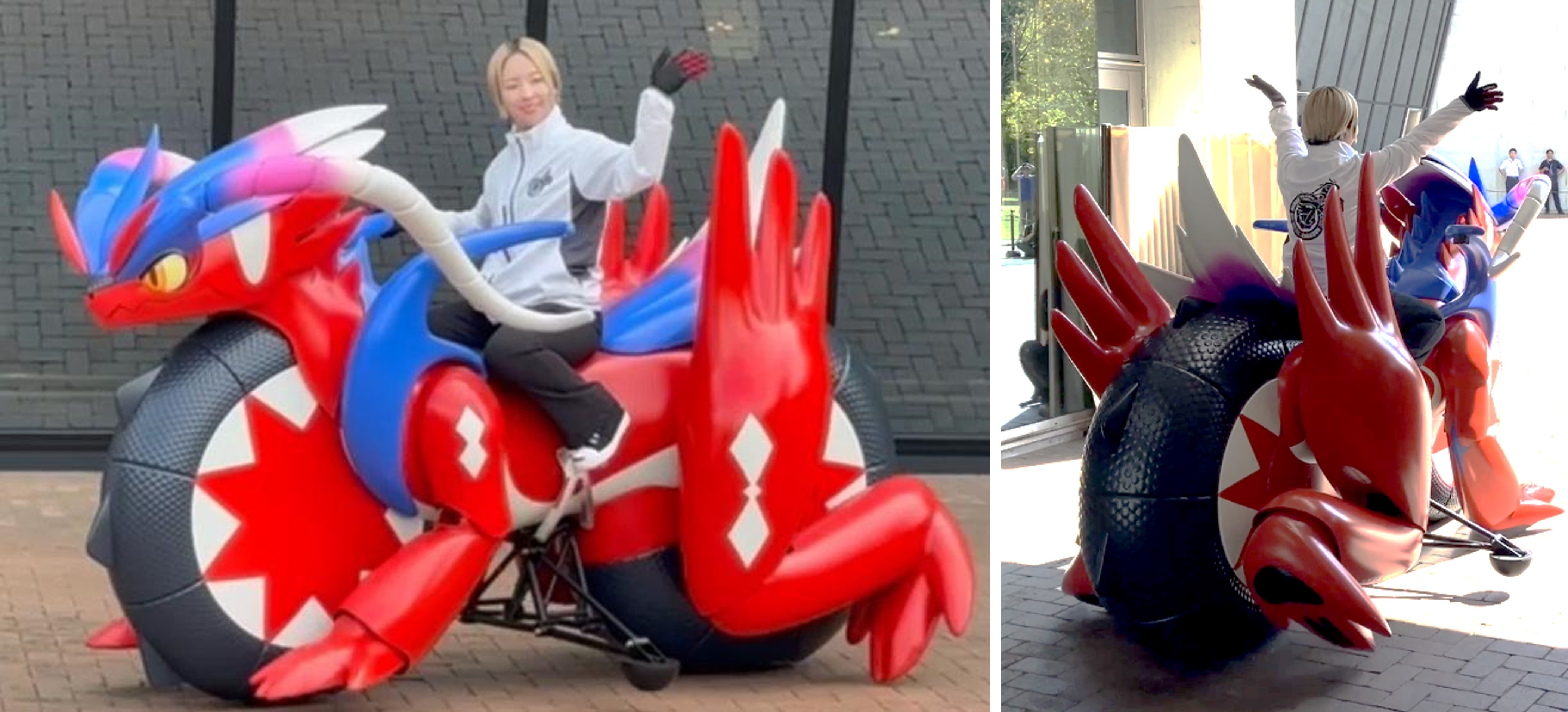}
\caption{Snapshot of outdoor experimental scene.  The robot is primarily controlled by the rider's body weight shifting, enabling hands-free maneuverability.}
\label{fig:test_scene}
\end{center}
\end{figure*}

\begin{figure}[tb]
\begin{center}
\includegraphics[width=80mm]{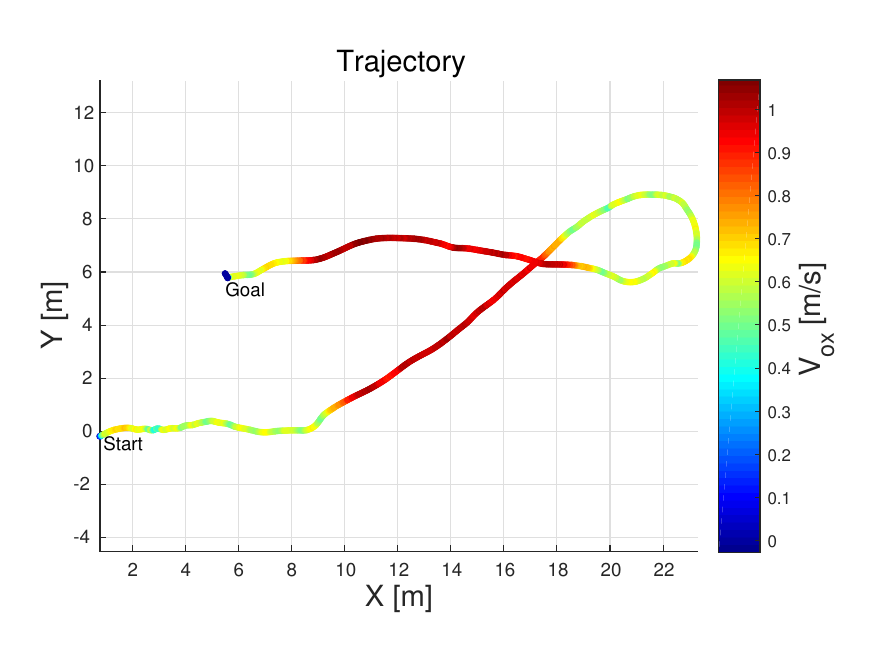}
\caption{Trajectory of the robot during outdoor riding tests. The trajectory is visualized based on odometry data.}
\label{fig:tra}
\end{center}
\end{figure}

The experimental scene of outdoor riding is shown in Fig.~\ref{fig:test_scene}.
The wheeled locomotion trajectory is shown in Fig.~\ref{fig:tra}.
The robot followed a trajectory combining straight-line and turning motions with the straight segments executed at higher speeds to demonstrate the gallop motion.
\par
\begin{figure}[tb]
\begin{center}
\includegraphics[width=80mm]{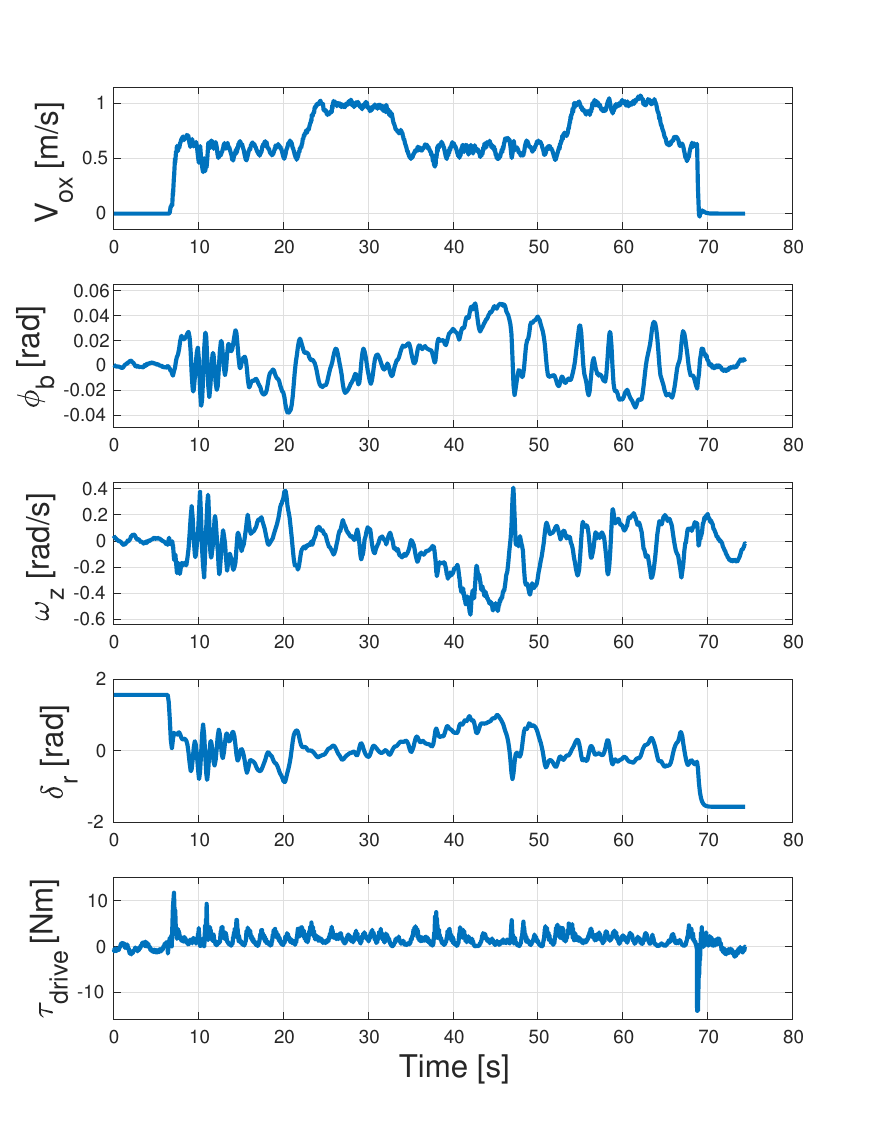}
\caption{Results of outdoor riding tests focusing on robot body dynamics. }
\label{fig:plot1}
\end{center}
\end{figure}

 Subsequently, the robot dynamics variables during wheeled locomotion are shown in Fig.~\ref{fig:plot1}.
For safety, the robot is operated with external outriggers to ensure that $|\phi_{b}|$ does not exceed 0.088 rad. From the graph of $\phi_{b}$
, it can be observed that the robot is able to travel without the outriggers making ground contact. The incorporation of the VEF structure suppresses instability at the onset of running, preventing falls and enabling robust wheeled locomotion. Furthermore, from the graph of $\delta_{r}$
, it is evident that the robot maintains self-balancing at $\delta_{r} = \pm 0.5\pi$ rad while standstill, and during wheeled locomotion it achieves stable operation by controlling $\delta_{r}$. 
Furthermore, from the graphs of $V_{ox}$ and $\tau_{drive}$, it can be observed that the robot is capable of wheeled locomotion while maintaining stability under fore-aft perturbations. 
\par
\begin{figure*}[tb]
\begin{center}
\includegraphics[width=190mm]{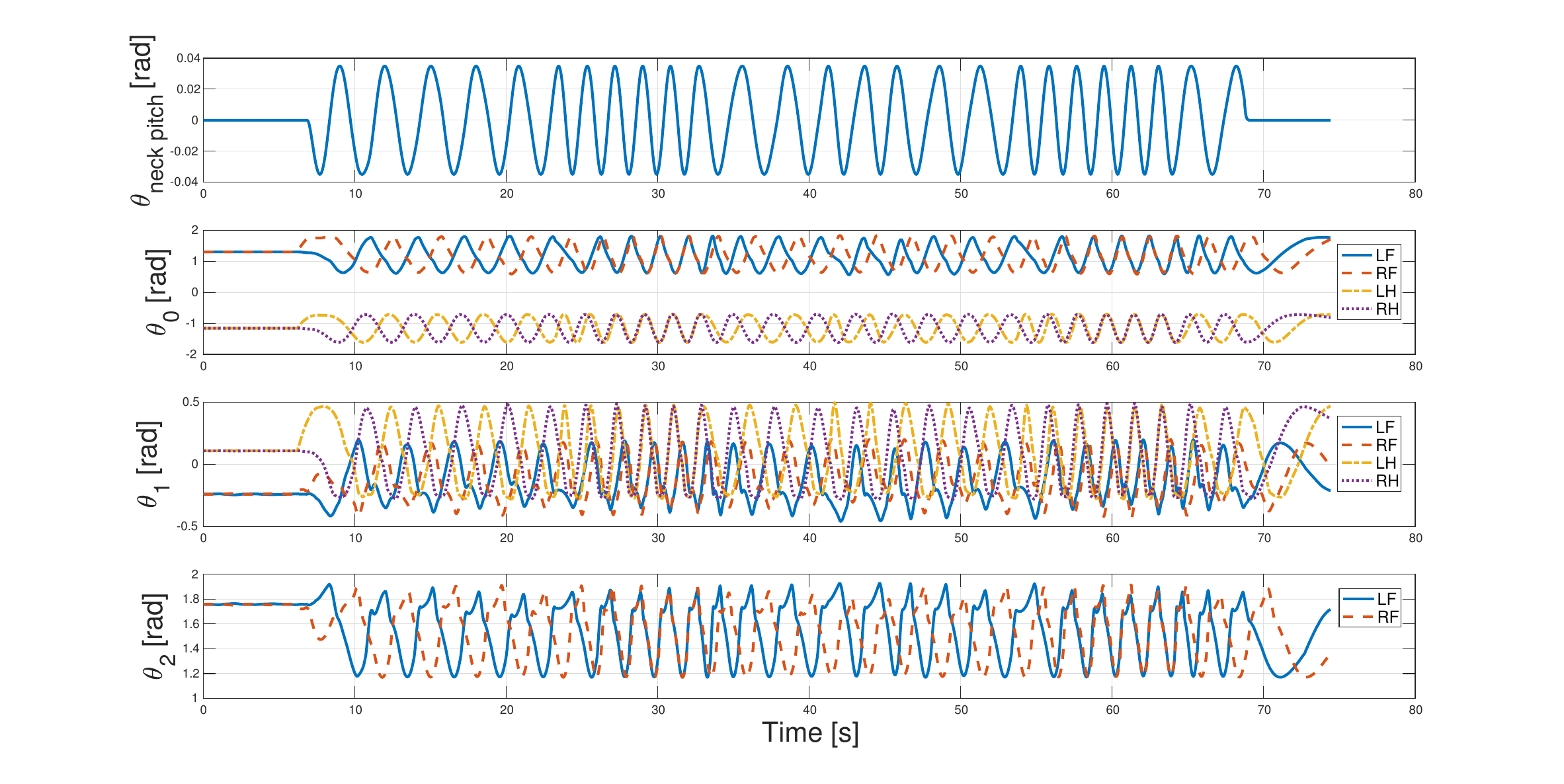}
\caption{Results of outdoor riding tests for the neck and limbs. }
\label{fig:plot3}
\end{center}
\end{figure*}

\begin{figure*}[tb]
\begin{center}
\includegraphics[width=160mm]{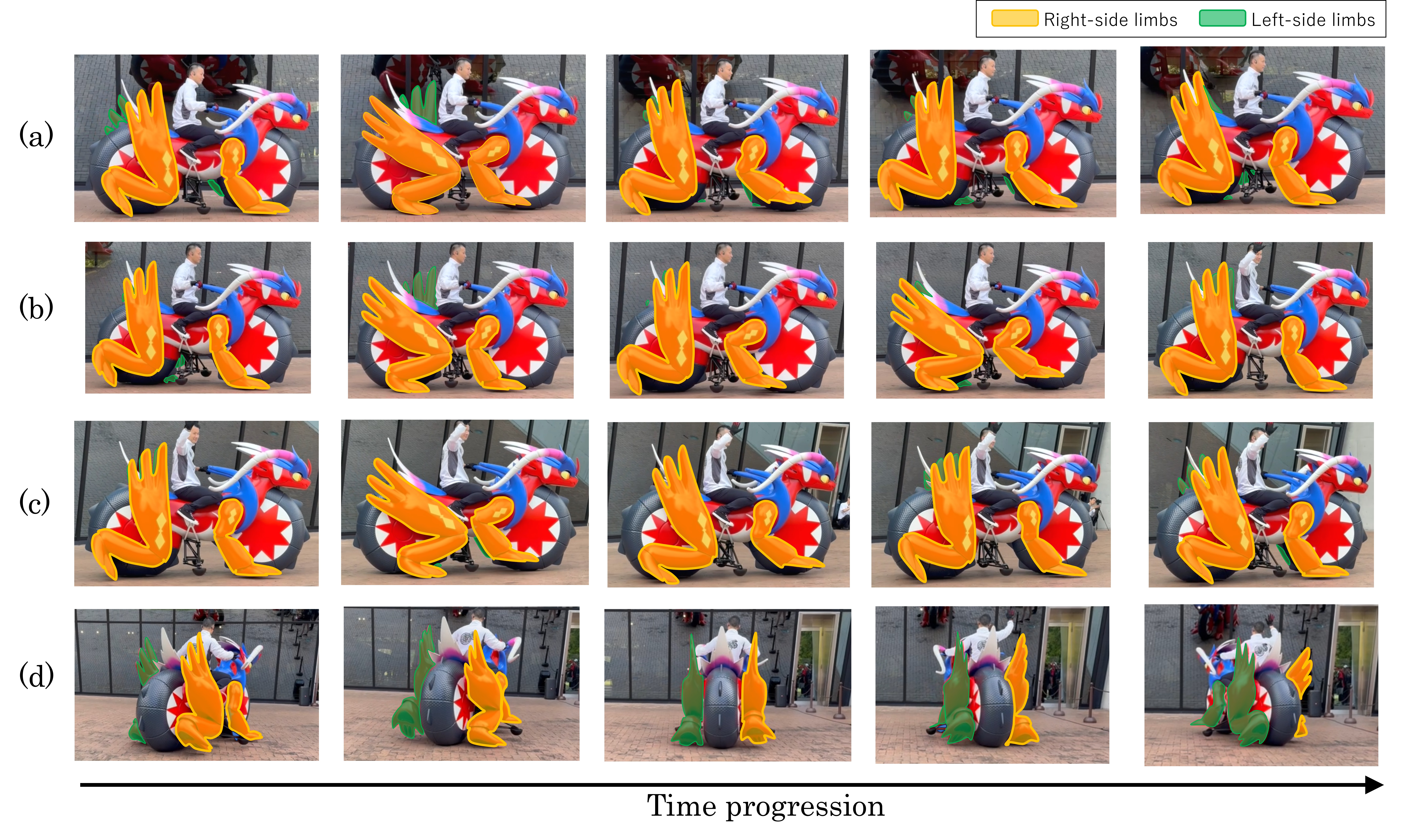}
\caption{Time-series visualization of each gait pattern: (a) Trot, (b) Transition from trot to gallop, (c) Gallop, (d) Cornering. (a) shows that a trot gait, in which diagonal pairs of limbs move simultaneously, is successfully achieved. In addition, (c) shows that a gallop gait, characterized by a sequential motion of the fore and hind limbs, is achieved.}
\label{fig:gait_results}
\end{center}
\end{figure*}

 Next, the results of neck and limb movements during wheeled locomotion are shown in Fig.~\ref{fig:plot3}.
From the graph, it can be observed that during the periods from 22 s to 34 s and from 52 s to 62 s, when the wheeled locomotion speed was increased, the left and right limbs moved synchronously, realizing a gallop gait. Outside of these periods, a trot gait was achieved, with diagonal limbs moving in sync. Furthermore, the top graph shows that the neck movement cycle varies according to the wheeled locomotion speed.
The summary of the above results as time-series images is shown in Fig.~\ref{fig:gait_results}.
 \par
\begin{figure}[tb]
\begin{center}
\includegraphics[width=80mm]{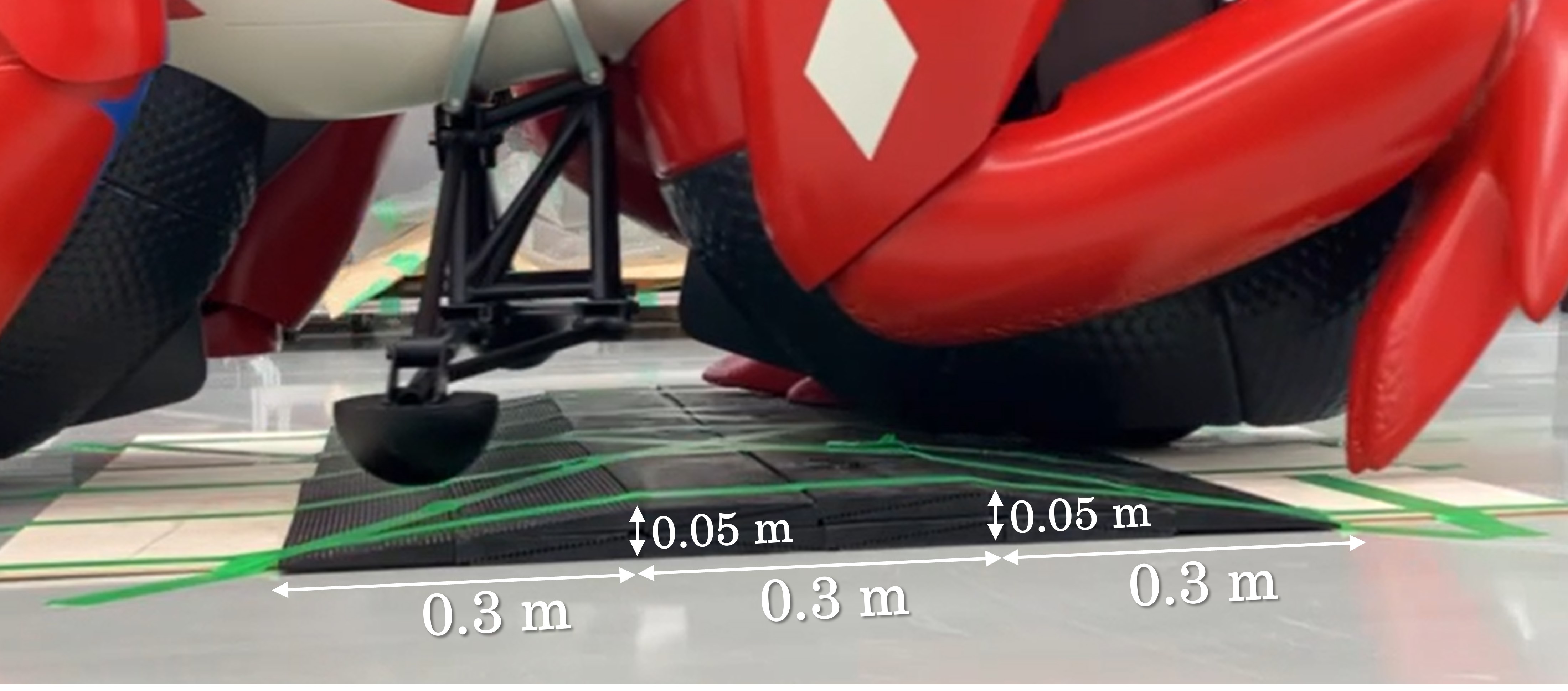}
\caption{Step-over test scene with a cable ramp model during straight motion. }
\label{fig:robust_locomotion}
\end{center}
\end{figure}

\begin{figure}[tb]
\begin{center}
\includegraphics[width=80mm]{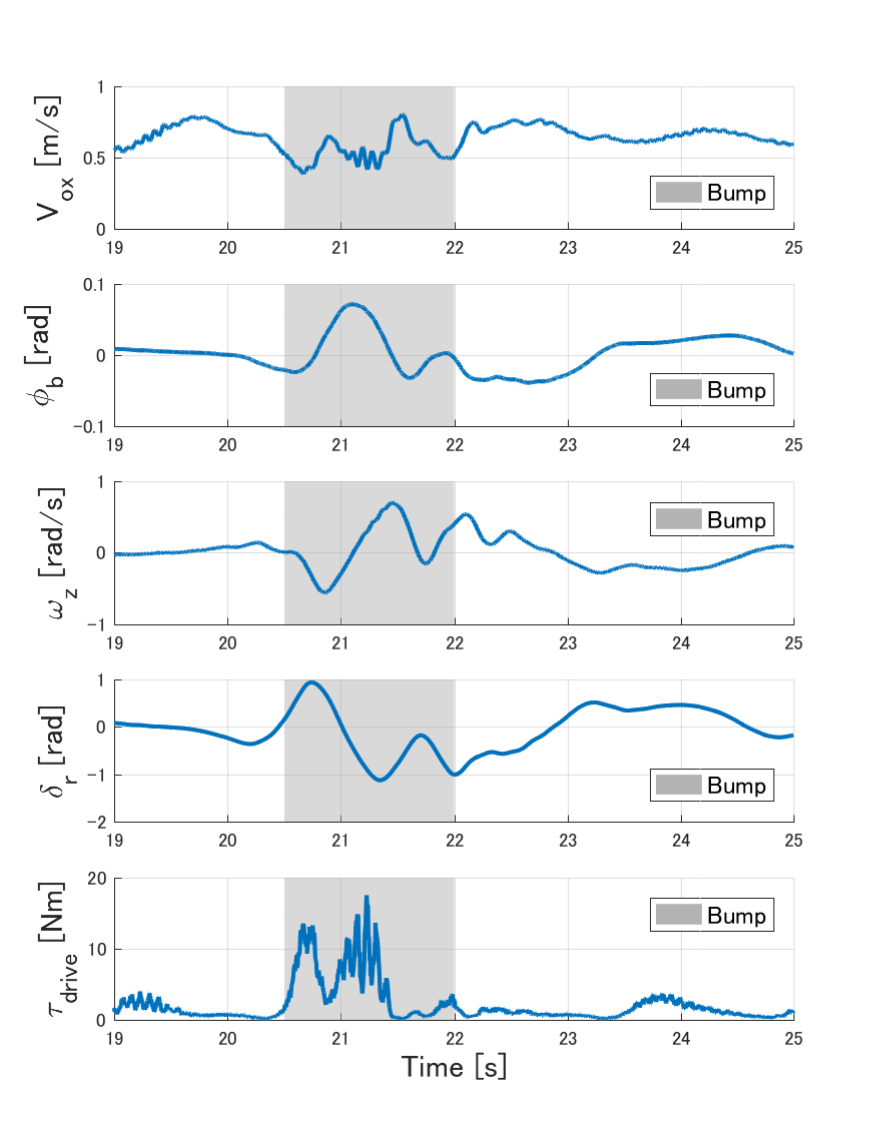}
\caption{Results of step-over test while moving straight ahead focusing on robot body dynamics. }
\label{fig:bump}
\end{center}
\end{figure}

 Considering the use in public outdoor events, it is desirable for the robot to possess sufficient robustness to traverse raised obstacles, such as cable ramps. Therefore, a step‑over test was conducted as shown in Fig.~\ref{fig:robust_locomotion}. 
 As illustrated in Fig.~\ref{fig:bump}, although slight lateral sway occurred when negotiating the obstacle, the robot was able to traverse it robustly.

\section{Discussion}
\par By employing the VEF framework, it was possible to enhance robustness while improving maneuverability since the desired state generated by the system side converges toward the state output resulting from the rider's control. This system-side adaptation is also a key concept in the development of partner mobility and is expected to become an important approach in future mobility control.
\par In the proposed rear-wheel steering robot, as shown in Fig.~\ref{fig:gait_results}(d),  the rider must shift their weight in the opposite direction during turning compared to conventional two-wheeled vehicles. Therefore, riders who are accustomed to bicycles or motorcycles and are operating this robot for the first time may require a certain level of familiarization. 
Furthermore, it remains a topic for future work to quantitatively evaluate the influence of the rider on the control system and the ease of operability. 
\par To prioritize straight-line maneuverability, the augmented system model in (\ref{state_eq_run}) was constructed. As a result, as shown in (\ref{K_P}), unstable pole-zero cancellation occurs at the wheeled locomotion speed where $V_{ox} = L_{f}\omega$, leading to a speed constraint of $V_{ox} < L_{f}\omega$. In this study, the speed at which the limb motors become overloaded, denoted as $V_{ox_{limb}}$, was found to satisfy $V_{ox_{limb}} < L_{f}\omega$, indicating no practical issues. 
While this depends on the wheeled locomotion mode, operating continuously at speeds exceeding 2.2 m/s results in a high load on the limb motors; therefore, $V_{ox_{limb}}$ was set to 2.2 m/s.
However, if higher wheeled locomotion speeds are desired in future applications, improvements in both hardware and control models will be necessary.

\section{Conclusion}
In this study, we developed a two-wheeled animatronics robot equipped with limbs and neck, which can be freely operated while being ridden. To enable stable self balancing movement even when the limbs and neck move during riding, we proposed a robust control method incorporating a virtual external force loop. Furthermore, by transitioning the gait from trot to gallop according to the wheeled locomotion speed, and by generating fore-aft perturbations of the center of mass based on acceleration and deceleration synchronized with limb phases, natural motion as an animatronics robot was successfully achieved. 

\section*{ACKNOWLEDGMENT}
In this work, under the supervision of the Pok$\acute{{\mathrm{e}}}$mon Company, we created the legendary Pok$\acute{{\mathrm{e}}}$mon Koraidon from \textit{Pok$\acute{{\mathrm{e}}}$mon Scarlet}—\copyright  2026 Pok$\acute{{\mathrm{e}}}$mon; \copyright  1995–2026 Nintendo/Creatures Inc./GAME FREAK inc.—as Honda Koraidon by integrating motorcycle engineering with robotics technology.


\end{document}